\documentclass[runningheads]{llncs}
\usepackage{graphicx}
\usepackage{epstopdf}
\usepackage{color}
\usepackage{cite}
\usepackage{amsfonts, amssymb, amsmath}
\usepackage{multirow}
\usepackage{epsf}
\usepackage{longtable}
\usepackage{booktabs}
\usepackage[table]{xcolor}
\usepackage{ragged2e}
\usepackage{subfigure}
\usepackage{marvosym}
\usepackage{rotating}
\usepackage{hyperref}
\hypersetup{hidelinks,
	colorlinks=true,
	allcolors=black,
	pdfstartview=Fit,
	breaklinks=true}

\begin{document}

\title{Uncertainty-informed Mutual Learning for Joint Medical Image Classification and Segmentation}
\author{Kai Ren\inst{1,2,*}, Ke Zou\inst{1,2,*}, Xianjie Liu\inst{2}, Yidi Chen\inst{3}, Xuedong Yuan\inst{1,2}\textsuperscript{\Letter},\\ Xiaojing Shen\inst{1,4}, Meng Wang\inst{5}, Huazhu Fu\inst{5}\textsuperscript{\Letter}}
\authorrunning{K. Ren et al.}
\institute{National Key Laboratory of Fundamental Science on Synthetic Vision,\\ Sichuan University, Sichuan, China \and College of Computer Science, Sichuan University, Sichuan, China \and Department of Radiology, West China Hospital, Sichuan University, Sichuan, China \and College of Mathematics, Sichuan University, Sichuan, China \and Institute of High Performance Computing, Agency for Science, Technology and Research, Singapore\\
\email{yxd@scu.edu.cn;hzfu@ieee.org}}
\renewcommand{\thefootnote}{}
\footnotetext{* denotes equal contribution.}
\maketitle

\begin{abstract}
Classification and segmentation are crucial in medical image analysis as they enable accurate diagnosis and disease monitoring. However, current methods often prioritize the mutual learning features and shared model parameters, while neglecting the reliability of features and performances. In this paper, we propose a novel Uncertainty-informed Mutual Learning (UML) framework for reliable and interpretable medical image analysis. Our UML introduces reliability to joint classification and segmentation tasks, leveraging mutual learning with uncertainty to improve performance. To achieve this, we first use evidential deep learning to provide image-level and pixel-wise confidences. Then, an uncertainty navigator is constructed for better using mutual features and generating segmentation results. Besides, an uncertainty instructor is proposed to screen reliable masks for classification. Overall, UML could produce confidence estimation in features and performance for each link (classification and segmentation). The experiments on the public datasets demonstrate that our UML outperforms existing methods in terms of both accuracy and robustness. Our UML has the potential to explore the development of more reliable and explainable medical image analysis models. 
\end{abstract}

\section{Introduction}
Accurate and robust classification and segmentation of the medical image are powerful tools to inform diagnostic schemes. In clinical practice, the image-level classification and pixel-wise segmentation tasks are not independent~\cite{zhou2021multi,kang2022thyroid}. Joint classification and segmentation can not only provide clinicians with results for both tasks simultaneously, but also extract valuable information and improve performance. However, improving the reliability and interpretability of medical image analysis is still reaching.

Considering the close correlation between the classification and segmentation, many researchers~\cite{harouni2018universal, kang2022thyroid, thomas2021interpretable, wang2023information, wang2021joint, zhou2021multi, zhu2021dsi} proposed to collaboratively analyze the two tasks with the help of sharing model parameters or task interacting. Most of the methods are based on sharing model parameters, which improves the performance by fully utilizing the supervision from multiple tasks~\cite{kang2022thyroid, zhou2021multi}. For example, Thomas \textit{et al.}~\cite{thomas2021interpretable} combined whole image classification and segmentation of skin cancer using a shared encoder. Task interacting is also a widely used method~\cite{mehta2018net, wang2021joint, zhu2021dsi} as it can introduce the high-level features and results produced by one task to benignly guide another. However, there has been relatively little research on introducing reliability into joint classification and segmentation. The reliability and interpretability of the model are particularly important for clinical tasks, a single result of the most likely hypothesis without any clues about how to make the decision might lead to misdiagnoses and sub-optimal treatment~\cite{kohl2018probabilistic, wang2023information}. One potential way of improving reliability is to introduce uncertainty for the medical image analysis model. 

The current uncertainty estimation method can roughly include the Dropout-based~\cite{lakshminarayanan2017simple}, ensemble-based~\cite{gal2016dropout, smith2018understanding, srivastava2014dropout}, deterministic-based methods~\cite{van2020uncertainty} and evidential deep learning~\cite{han2021trusted, zou2022tbrats, sensoy2018evidential, zou2023evidencecap, wang2023uncertainty}. All of these methods are widely utilized in classification and segmentation applications for medical image analysis. Abdar \textit{et al.}~\cite{abdar2021uncertainty} employed three uncertainty quantification methods (Monte Carlo dropout, Ensemble MC dropout, and Deep Ensemble) simultaneously to deal with uncertainty estimation during skin cancer image classification. Zou \textit{et al.}~\cite{zou2022tbrats} proposed TBraTS based on evidential deep learning to generate robust segmentation results for brain tumor and reliable uncertainty estimations. Unlike the aforementioned methods, which only focus on uncertainty in either medical image classification or segmentation. Furthermore, none of the existing methods have considered how pixel-wise and image-level uncertainty can help improve performance and reliability in mutual learning.

Based on the analysis presented above, we design a novel Uncertainty-informed Mutual Learning (UML) network for medical image analysis in this study. Our UML not only enhances the image-level and pixel-wise reliability of medical image classification and segmentation, but also leverages mutual learning under uncertainty to improve performance. Specifically, we adopt evidential deep learning~\cite{sensoy2018evidential,zou2022tbrats} to simultaneously estimate the uncertainty of both to estimate image-level and pixel-wise uncertainty. We introduce an Uncertainty Navigator for segmentation (UN) to generate preliminary segmentation results, taking into account the uncertainty of mutual learning features. We also propose an Uncertainty Instructor for classification (UI) to screen reliable masks for classification based on the preliminary segmentation results. Our UML represents pioneering work in introducing reliability and interpretability to joint classification and segmentation, which has the potential to the development of more trusted medical analysis tools. \footnote{Our code has been released in \href{https://github.com/KarryRen/UML}{https://github.com/KarryRen/UML}.}

\section{Method}
\begin{figure}[!t]
\centering
\includegraphics[width=0.9\linewidth]{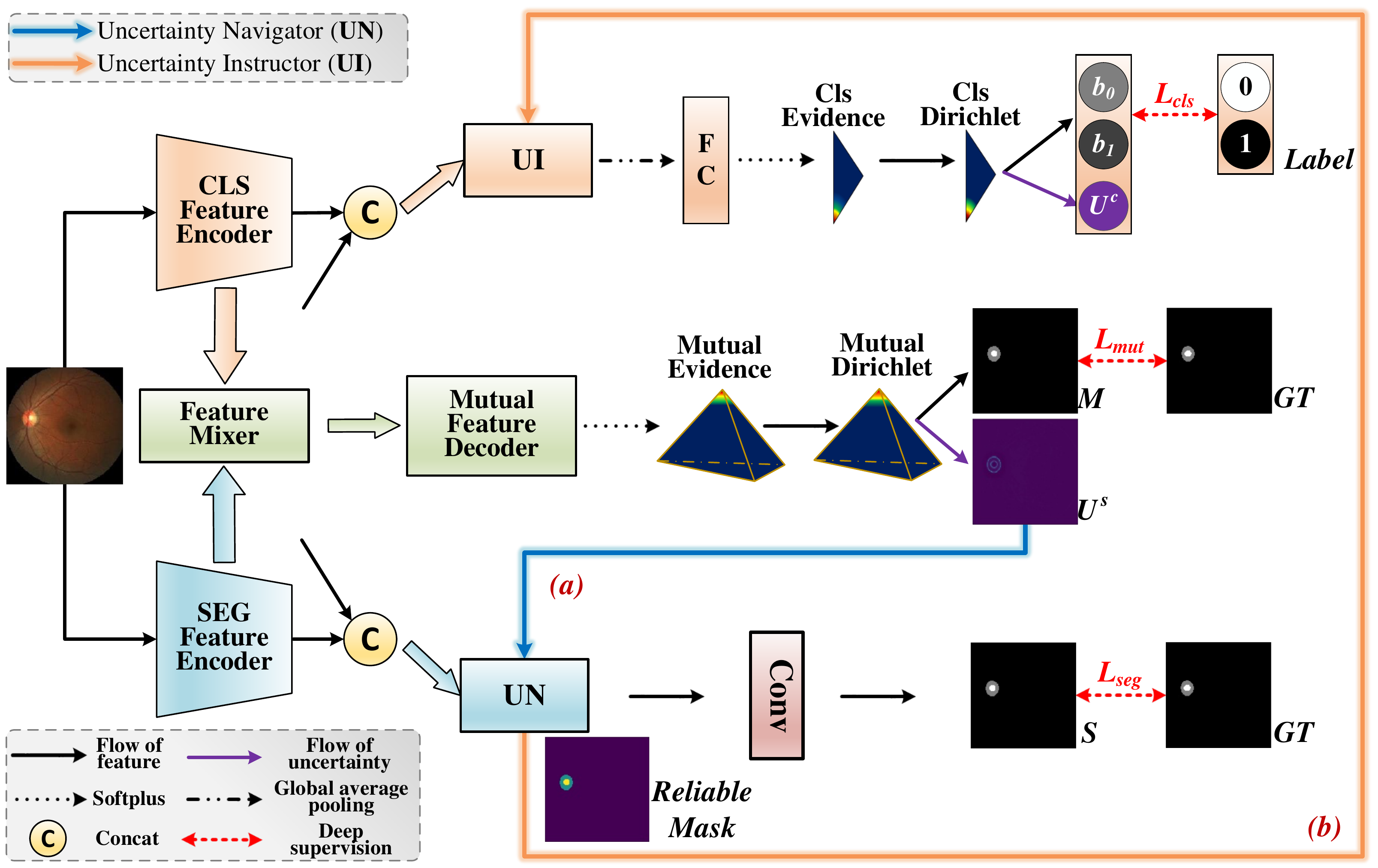}
\caption{The framework of Uncertainty-informed Mutual Learning network.}
\label{F_1}
\end{figure}

The overall architecture of the proposed UML, which leverages mutual learning under uncertainty, is illustrated in Fig.~\ref{F_1}. Firstly, Uncertainty Estimation for Classification and Segmentation adapts evidential deep learning to provide image-level and pixel-wise uncertainty. Then, Trusted Mutual Learning not only utilizes the proposed UN to fully exploit pixel-wise uncertainty as the guidance for segmentation but also introduces the UI to filter the feature flow between task interaction. 

Given an input medical image $I, I\in \mathbb{R}^{H, W}$, where $H, W$ are the height and width of the image, separately. To maximize the extraction of specific information required for two different tasks while adequately mingling the common feature which is helpful for both classification and segmentation, $I$ is firstly fed into the dual backbone network that outputs the classification feature maps $f^{c}_i, i\in1, ..., 4$ and segmentation feature maps $f^{s}_i, i\in1, ..., 4$, where $i$ denotes the $i^{th}$ layer of the backbone. Then following~\cite{zou2022interactive}, we construct the Feature Mixer using Pairwise Channel Map Interaction to mix the original feature and get the mutual feature maps $f^{m}_i, i\in1, ..., 4$. Finally, we combine the last layer of mutual feature with the original feature.

\subsection{Uncertainty Estimation for Classification and Segmentation}
\textbf{Classification Uncertainty Estimation.} For the $K$ classification problems, we utilize Subjective Logic~\cite{jsang2016subjective} to produce the belief mass of each class and the uncertainty mass of the whole image based on \textit{evidence}. Accordingly, given a classification result, its $K + 1$ mass values are all non-negative and their sum is one:
\begin{equation}
\label{E_p1}
{\sum_{k=1}^K b^c_k + U^c = 1,}
\end{equation}
where $b^c_k \ge 0$ and $U^c \ge 0$ denote the probability belonging to the $k^{th}$ class and the overall uncertainty value, respectively. As shown in Fig.~\ref{F_1}, the \textit{cls evidence} $e^{c} = [e^{c}_1, \dots, e^{c}_K]$ for the classification result is acquired by an activation function layer softplus and $e^{c}_k \ge 0$. Then the \textit{cls Dirichlet} distribution can be parameterize by $\alpha^{c} = [\alpha^{c}_1, \dots, \alpha^{c}_K]$, which associated with the \textit{cls evidence} $e^{c}_k$, \textit{i.e.} $\alpha^{c}_k=e^{c}_k+1$. In the end, the image-level belief mass and the uncertainty mass of the classification can be calculated by 
\begin{equation}
\label{E_2.2}
{b^{c}_k=\frac{e^{c}_k}{T^{c}}=\frac{\alpha^{c}_k-1}{T^{c}},\  \text{and}\  \ U^{c}=\frac{K}{T^{c}},}
\end{equation}
where $T^{c}=\sum_{k=1}^K{\alpha^{c}_k}=\sum_{k=1}^K{(e^{c}_k+1)}$ represents the Dirichlet strength. Actually, Eq.~\ref{E_2.2} describes such a phenomenon that the higher the probability assigned to the $k^{th}$ class, the more evidence observed for $k^{th}$ category should be. 

\textbf{Segmentation Uncertainty Estimation.} Essentially, segmentation is the classification for each pixel of a medical image. Given a pixel-wise segmentation result, following~\cite{zou2022tbrats} the \textit{seg Dirichlet} distribution can be parameterized by $\alpha^{s(h, w)} = [\alpha^{s(h, w)}_1, \dots, \alpha^{s(h, w)}_Q], (h, w)\in(H, W)$. We can compute the belief mass and uncertainty mass of the input image by
\begin{equation}
\label{E_2.4}
{b^{s(h, w)}_q=\frac{e^{s(h, w)}_q}{T^{s(h, w)}}=\frac{\alpha^{s(h, w)}_q-1}{T^{s(h, w)}},\ \  \text{and} \ \ u^{s(h, w)}=\frac{Q}{T^{s(h, w)}},}
\end{equation}
where ${b_q^{s(h, w)}} \ge 0$ and $u^{s(h, w)} \ge 0$ denote the probability of the pixel at coordinate $(h, w)$ for the $q^{th}$ class and the overall uncertainty value respectively. We also define $U^{s}=\{u^{s(h, w)}, (h, w)\in(H, W)\}$ as the pixel-wise uncertainty of the segmentation result. 

\subsection{Uncertainty-informed Mutual Learning}
\begin{figure}[!t]
\centering
\includegraphics[width=0.7\linewidth]{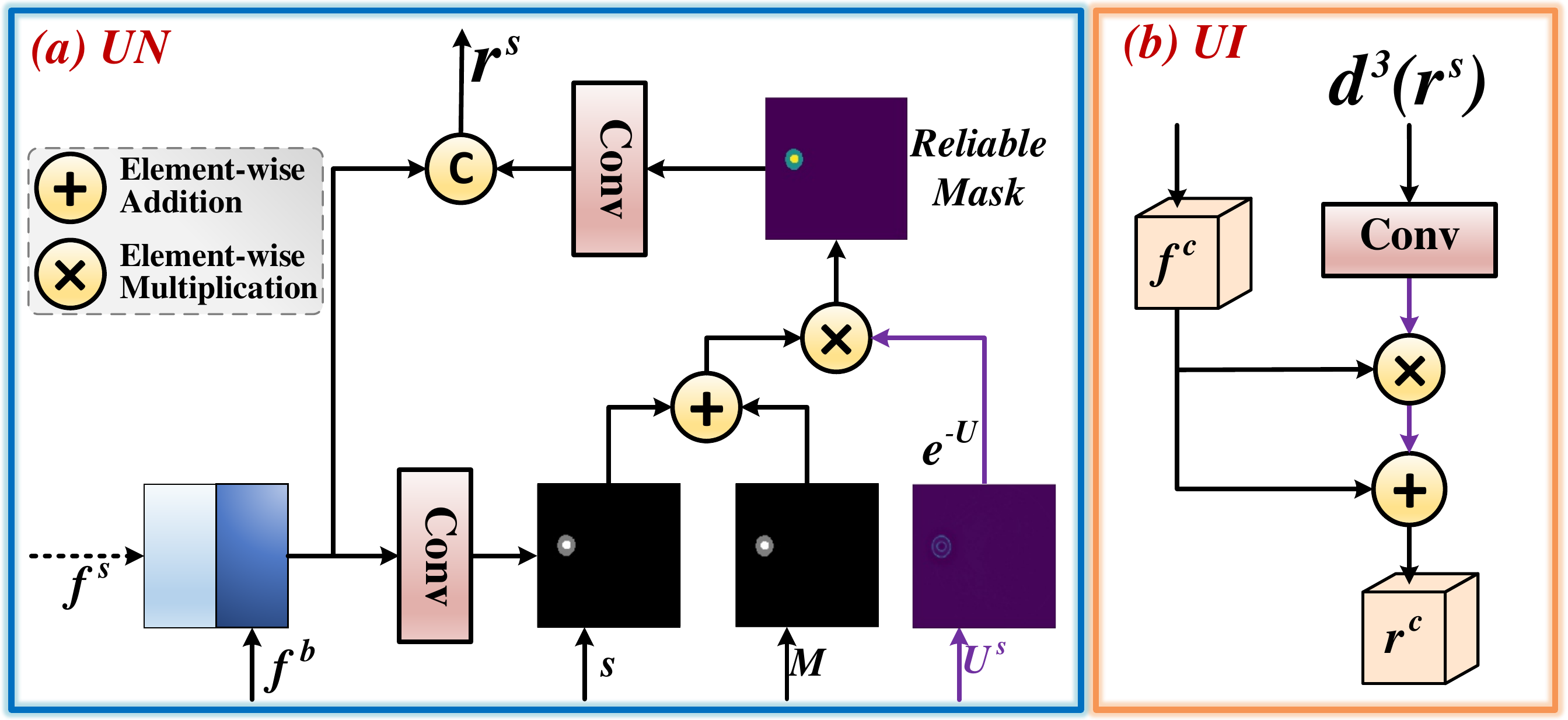}
\caption{Details of (a) Uncertainty Navigator (UN) and (b) Uncertainty Instructor 
(UI).}
\label{F_2}
\end{figure}

\textbf{Uncertainty Navigator for Segmentation.} Actually, we have already obtained an initial segmentation mask $M={\alpha^{s}}, $ $M\in(Q, H, W)$ through estimating segmentation uncertainty, and achieved lots of valuable features such as lesion location. In our method, appropriate uncertainty guided decoding on the feature list can obtain more reliable information and improve the performance of segmentation~\cite{cui2023uncertainty,kim2021uacanet,zhang2022preynet}. So we introduce Uncertainty Navigator for Segmentation(UN) as a feature decoder, which incorporates the pixel-wise uncertainty in $U^{s}$and lesion location information in $M$ with the segmentation feature maps to generate the segmentation result and reliable features.
Having a UNet-like architecture~\cite{ronneberger2015u}, UN computes segmentation $s_i, i\in1, .., 4$ at each layer, as well as introduces the uncertainty in the bottom and top layer by the same way. Take the top layer as an example, as shown in Fig.~\ref{F_2} (a), UN calculates the reliable mask $M^r$ by:
\begin{equation}
\label{E_3.11}
{M^r=(s_1\oplus M)\otimes e^{-U^{s}},}
\end{equation}
Then, the reliable segmentation feature $r^s$, which combines the trusted information in $M^r$ with the original features, is generated by: 
\begin{equation}
\label{E_3.1}
{r^s=Cat(Conv(M^r), Cat(f^{s}_1, f^{b}_2)),}
\end{equation}
where $f^s_1$ derives from jump connecting and $f_2^b$ is the feature of the $s_2$ with one up-sample operation. $Conv(\cdot)$ represents the convolutional operation, $Cat(\cdot, \cdot)$ denotes the concatenation. Especially, the $U^s$ is also used to guide the bottom feature with the dot product. The $r^s$ is calculated from the segmentation result $s_1$ and contains uncertainty navigated information not found in $s_1$. 

\textbf{Uncertainty Instructor for Classification.} In order to mine the complementary knowledge of segmentation as the instruction for the classification and eliminate intrusive features, we devise an Uncertainty Instructor for classification (UI) following~\cite{wang2023information}. Fig.~\ref{F_2} (b) shows the architecture of UI. It firstly generates reliable classification features $r^c$ fusing the initial classification feature maps $f^{c}_{4}$ and the rich information (e.g., lesion location and boundary characteristic) in $r^s$, which can be expressed by:
\begin{equation}
\label{E_3.3}
{r^c=f^{c}_{4} \oplus (Conv(d^{3}(r^s)) \otimes f^{c}_4),}
\end{equation}
where $d^n(\cdot)$ denotes that the frequency of down-sampling operations is $n$. Then the produced features are transformed into a semantic feature vector by the global average pooling. The obtained vector is converted into the final result (belief values) of classification with uncertainty estimation.

\subsection{Mutual Learning Process}
In a word, to obtain the final results of classification and segmentation, we construct an end-to-end mutual learning process, which is  supervised by a joint loss function. To obtain an initial segmentation result $M$ and a pixel-wise uncertainty estimation $U^s$, following~\cite{zou2022tbrats}, a mutual loss is used as:
\begin{equation}
\label{E_11}
{\mathcal{L}_{m}(\alpha^s, y^s) = \mathcal{L}_{ice}(\alpha^s, y^s) + \lambda^m_1\mathcal{L}_{KL}(\alpha^s) + \lambda^m_2\mathcal{L}_{Dice}(\alpha^s, y^s)}, 
\end{equation}
where $y^s$ is the Ground Truth (GT) of the segmentation. The hyperparameters $\lambda^m_1$ and $\lambda^m_2$ play a crucial role in controlling the Kullback-Leibler divergence (KL) and Dice score, as supported by~\cite{zou2022tbrats}. Similarly, in order to estimate the image-level uncertainty and classification results. a classification loss is constructed following~\cite{han2021trusted}, as: 
\begin{equation}
\label{E_15}
{\mathcal{L}_{c}(\alpha^c, y^c) = \mathcal{L}_{ace}(\alpha^c, y^c) + \lambda^c\mathcal{L}_{KL}(\alpha^c)}, 
\end{equation}
where $y^c$ is the true class of the input image. The hyperparameter $\lambda^c$ serves as a crucial hyperparameter governing the KL, aligning with previous work~\cite{han2021trusted}. To obtain reliable segmentation results, we also adopt deep supervision for the final segmentation result $S = \{s_i, i=1, ..., 4\}$, which can be denoted as: 
\begin{equation}
\label{E_16}
\mathcal{L}_{s} = \frac{\sum^4_{i=1}\mathcal{L}_{Dice}(\upsilon^{i-1}(s_i), y^s)}{4}, 
\end{equation}
where $\upsilon^{n}$ indicates the number of up-sampling is $2^{n}$. Thus, the overall loss function of our UML can be given as:
\begin{equation}
\label{E_18}
\mathcal{L}_{UML}(\alpha^s, \alpha^c, y^s, y^c) = w^m\mathcal{L}_{m}(\alpha^s, y^s) + w^c\mathcal{L}_{c}(\alpha^c, y^c) + w^s\mathcal{L}_{s},
\end{equation}
where $w^m, w^c, w^s$ denote the weights and are set $0.1, 0.5, 0.4$, separately.

\section{Experiments}

\noindent \textbf{Dataset \& Implementation.} We evaluate the our UML network on two datasets REFUGE~\cite{orlando2020refuge} and ISPY-1~\cite{newitt2016multi}. REFUGE contains two tasks, classification of glaucoma and segmentation of optic disc/cup in fundus images. The overall 1200 images were equally divided for training, validation, and testing. All images are uniformly adjusted to 256$\times$256 pixels. The tasks of ISPY-1 are the pCR prediction and the breast tumor segmentation. A total of 157 patients who suffer the breast cancer are considered - 43 achieve pCR and 114 non-pCR. For each case, we cut out the slices in the 3D image and totally got 1,570 2D images, which are randomly divided into the train, validation, and test datasets with 1,230, 170, and 170 slices, respectively. 

We implement the proposed method via PyTorch and train it on NVIDIA GeForce RTX 2080Ti. The Adam optimizer is adopted to update the overall parameters with an initial learning rate 0.0001 for 100 epochs. The scale of the regularizer is set as $1\times10^{-5}$. We choose VGG-16 and Res2Net as the encoders for classification and segmentation, separately.

\textbf{Compared Methods \& Metrics.} We compared our method with single-task methods and multi-task methods. (1) Single-task methods: (a) EC~\cite{evidential18}, (b) TBraTS~\cite{zou2022tbrats} and (c) TransUNet~\cite{chen2021transunet}. Evidential deep learning for classification (EC) first proposed to parameterize classification probabilities as Dirichlet distributions to explain evidence. TBraTS then extended EC to medical image segmentation. Meriting both Transformers and U-Net, TransUNet is a strong model for medical image segmentation. (2) Multi-task methods: (d) BCS~\cite{yang2017novel} and (e) DSI~\cite{zhu2021dsi}. The baseline of the Joint Classification and Segmentation framework (BCS) is a simple but useful way to share model parameters, which utilize two different encoders and decoders for learning respectively. The Deep Synergistic Interaction Network (DSI) has demonstrated superior performance in joint task. We adopt overall Accuracy (ACC) and F1 score (F1) as the evaluation criteria for the classification task. Dice score (DI) and Average Symmetric Surface Distance (ASSD) are chosen for the segmentation task.

\begin{table}[!t]
  \centering
  \caption{Evaluation of the classification and segmentation performance. The top-2 results are highlighted in bold and underlined ($p \leq 0.01$).}
  \resizebox{\linewidth}{!}{
    \begin{tabular}{c|ccccccccccccccc}
    \toprule
    \multicolumn{2}{c}{\multirow{3}[6]{*}{Method}} &       & \multicolumn{7}{c}{REFUGE}                            &       & \multicolumn{5}{c}{ISPY-1} \\
\cmidrule{4-10}\cmidrule{12-16}    \multicolumn{2}{c}{} &       & \multicolumn{2}{c}{CLS} &       & \multicolumn{4}{c}{SEG}       &       & \multicolumn{2}{c}{CLS} &       & \multicolumn{2}{c}{SEG} \\
\cmidrule{4-5}\cmidrule{7-10}\cmidrule{12-13}\cmidrule{15-16}    \multicolumn{2}{c}{} &       & $ACC$   & $F1$    &       & $DI_{disc}$ & $ASSD_{disc}$ & $DI_{cup}$ & $ASSD_{cup}$ &       & $ACC$   & $F1$    &       & $DI$    & $ASSD$ \\
\cmidrule{1-2}\cmidrule{4-5}\cmidrule{7-10}\cmidrule{12-13}\cmidrule{15-16}    \multicolumn{1}{c|}{\multirow{3}[2]{*}{Single-\newline{}task}} & EC    &       & 0.560  & 0.641  &       & \textbackslash{} & \textbackslash{} & \textbackslash{} & \textbackslash{} &       & 0.735  & 0.648  &       & \textbackslash{} & \textbackslash{} \\
          & TBraTS &       & \textbackslash{} & \textbackslash{} &       & 0.776  & 1.801  & 0.787  & 1.798  &       & \textbackslash{} & \textbackslash{} &       & 0.784  & 4.075  \\
          & TransUNet &       & \textbackslash{} & \textbackslash{} &       & 0.633  & 2.807  & 0.628  & 2.638  &       & \textbackslash{} & \textbackslash{} &       & 0.692  & 5.904  \\
\cmidrule{1-2}    \multicolumn{1}{c|}{\multirow{3}[2]{*}{Multi-\newline{}task}} & BCS   &       & 0.723  & 0.778  &       & \underline{0.802}  & \underline{1.692}  & \underline{0.831}  & \underline{1.532}  &       & \underline{0.758}  & \underline{0.692}  &       & 0.773  & \textbf{3.804} \\
          & DSI   &       & \underline{0.838}  & \underline{0.834}  &       & 0.793  & 2.030  & 0.811  & 1.684  &       & 0.741  & 0.673  &       & 0.760  & 4.165  \\
          & Ours  &       & \textbf{0.853} & \textbf{0.875} &       & \textbf{0.855} & \textbf{1.560} & \textbf{0.858} & \textbf{1.251} &       & \textbf{0.771} & \textbf{0.713} &       & \textbf{0.785} & \underline{3.927}  \\
    \bottomrule
    \end{tabular}%
    }
  \label{T_1}%
\end{table}%

\begin{figure}[!t]
\centering
\includegraphics[width=1\linewidth]{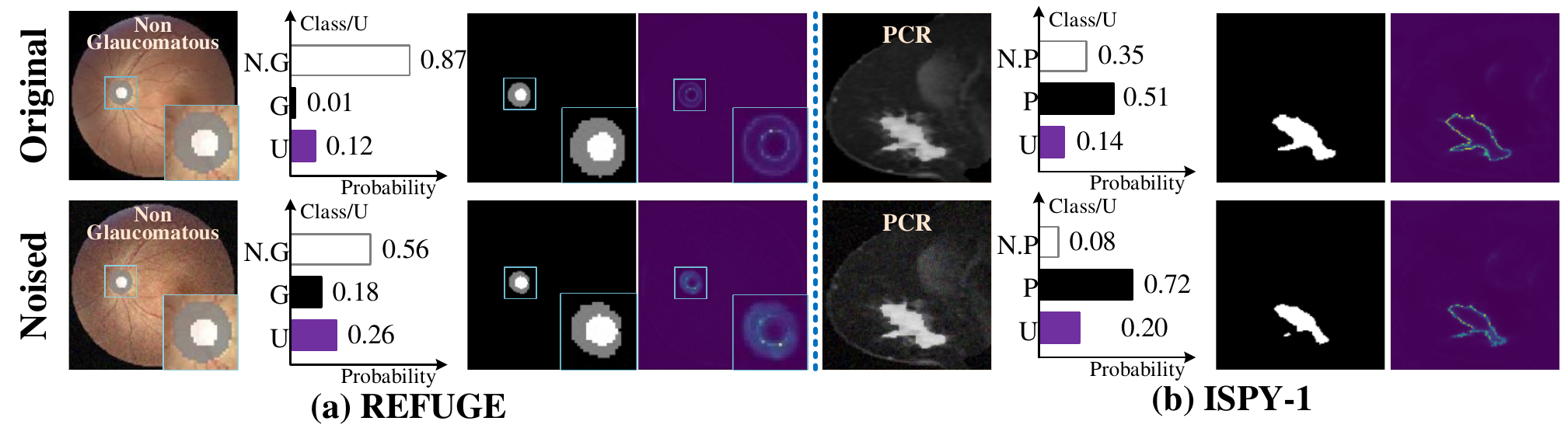}
\vskip -5pt
\caption{The visual result of segmentation and classification in REFUGE and ISPY-1. Top is the original image, and bottom is the input with Gaussian noise ($\sigma = 0.05$). From left to right, input (with GT), the result of classification (belief and image-level uncertainty), the result of segmentation, pixel-wise uncertainty.}
\label{F_3}
\end{figure}

\textbf{Comparison with single- and multi-task methods.} As shown in Table~\ref{T_1}, we report the performance on the two datasets of the proposed UML and other methods. By comparison, we can observe the fact that the accuracy of the model results is low if either classification or segmentation is done in isolation, the ACC has only just broken 0.5 in EC. But joint classification and segmentation changes this situation, the performance of BCS and DSI improves considerably, especially the ACC and the Dice score of optic cup. Excitingly, our UML not only achieves the best classification performance in ACC (85.3\%) and F1 (0.875) with significant increments of 1.8\%, 4.9\%, but also obtains the superior segmentation performance with increments of 6.6\% in $DI_{disc}$ and 3.2\% in $DI_{cup}$. A similar improvement can be observed in the experimental results in ISPY-1.

\begin{table}[!t]
  \centering
  \caption{The quantitative comparisons on the REFUGE dataset with vary NOISE levels.}
    \resizebox{\linewidth}{!}{
    \begin{tabular}{ccccccccccccccccc}
    \toprule
    \multirow{3}[6]{*}{REFUGE} &       & \multicolumn{7}{c}{0.030}                            &       & \multicolumn{7}{c}{0.050} \\
\cmidrule{3-9}\cmidrule{11-17}          &       & \multicolumn{2}{c}{CLS} &       & \multicolumn{4}{c}{SEG}       &       & \multicolumn{2}{c}{CLS} &       & \multicolumn{4}{c}{SEG} \\
\cmidrule{3-4}\cmidrule{6-9}\cmidrule{11-12}\cmidrule{14-17}          &       & $ACC$   & $F1$    &       & $DI_{disc}$ & $ASSD_{disc}$ & $DI_{cup}$ & $ASSD_{cup}$ &       & $ACC$   & $F1$    &       & $DI_{disc}$ & $ASSD_{disc}$ & $DI_{cup}$ & $ASSD_{cup}$ \\
\cmidrule{1-1}\cmidrule{3-4}\cmidrule{6-9}\cmidrule{11-12}\cmidrule{14-17}    BCS   &       & 0.620  & 0.694  &       & \underline{0.743}  & \underline{2.104}  & \underline{0.809}  & \underline{1.670}  &       & 0.430  & 0.510  &       & \underline{0.610}  & \underline{3.142}  & \underline{0.746}  & \underline{2.188}  \\
    DSI   &       & \underline{0.675}  & \underline{0.733}  &       & 0.563  & 9.196  & 0.544  & 9.705  &       & \underline{0.532}  & \underline{0.574}  &       & 0.409  & 8.481  & 0.364  & 9.794  \\
    Ours  &       & \textbf{0.827} & \textbf{0.857} &       & \textbf{0.830} & \textbf{1.752} & \textbf{0.840} & \textbf{1.407} &       & \textbf{0.733} & \textbf{0.785} &       & \textbf{0.744} & \textbf{2.142} & \textbf{0.778} & \textbf{2.009} \\
    \bottomrule
    \end{tabular}%
    }
  \label{T_2}%
\end{table}%

\textbf{Comparison under noisy data.} To further valid the reliability of our model, we introduce Gaussian noise with various levels of standard deviations ($\sigma$) to the input medical images. The comparison results are shown in Table~\ref{T_2}. As can be observed that, the accuracy of classification and segmentation significantly decreases after adding noise to the raw data. However, benefiting from the uncertainty-informed guiding, our UML consistently deliver impressive results. In Fig.~\ref{F_3}, we show the output of our model under the noise. It is obvious that both the image-level uncertainty and the pixel-wise uncertainty respond reasonably well to noise. These experimental results can verify the reliability and interpre of the uncertainty guided interaction between the classification and segmentation in the proposed UML. The results of more qualitative comparisons can be found in the Supplementary Material.

\textbf{Ablation Study.} As illustrated in Table~\ref{T_3}, both of the proposed UN and UI play important roles in trusted mutual learning. The baseline method is BCS. MD represents the mutual feature decoder. It is clear that the performance of classification and segmentation is significantly improved when we introduce supervision of mutual features. As we thought, the introduction of UN and UI takes the reliability of the model to a higher level.

\begin{table}[!t]
  \centering
  \caption{The result of Ablation Study.}
  \resizebox{0.7\linewidth}{!}{
    \begin{tabular}{ccccccccccc}
    \toprule
    \multicolumn{1}{c}{\multirow{2}[4]{*}{MD}} & \multirow{2}[4]{*}{UN} & \multirow{2}[4]{*}{UI} &       & \multicolumn{2}{c}{CLS} &       & \multicolumn{4}{c}{SEG} \\
    \cmidrule{5-6}\cmidrule{8-11}          &       &       &       & $ACC$   & $F1$    &       & $DI_{disc}$ & $ASSD_{disc}$ & $DI_{cup}$ & $ASSD_{cup}$ \\
    \cmidrule{1-3}\cmidrule{5-11}    $\checkmark$     &       &       &       & 0.765  & 0.810  &       & 0.835  & \underline{1.454}  & \underline{0.841}  & \underline{1.423}  \\
    $\checkmark$     & $\checkmark$     &       &       & 0.813  & 0.845  &       & \underline{0.836}  & \textbf{1.333} & 0.826  & 1.525  \\
    $\checkmark$     &       & $\checkmark$     &       & \underline{0.828}  & \underline{0.856}  &       & 0.786  & 1.853  & 0.823  & 1.593  \\
    $\checkmark$     & $\checkmark$     & $\checkmark$     &       & \textbf{0.853} & \textbf{0.875} &       & \textbf{0.855} & 1.560  & \textbf{0.858} & \textbf{1.251} \\
    \bottomrule
    \end{tabular}%
    }
  \label{T_3}%
\end{table}%

\section{Conclusion}
In this paper, we propose a novel deep learning approach, UML, for joint classification and segmentation of medical images. Our approach is designed to improve the reliability and interpretability of medical image classification and segmentation, by enhancing image-level and pixel-wise reliability estimated by evidential deep learning, and by leveraging mutual learning with the proposed UN and UI modules. Our extensive experiments demonstrate that UML outperforms baselines and introduces significant improvements in both classification and segmentation. Overall, our results highlight the potential of UML for enhancing the performance and interpretability of medical image analysis.

\small{\textbf{Acknowledgements:} This work was supported by the National Research Foundation, Singapore under its AI Singapore Programme (AISG Award No: AISG2-TC-2021-003), A*STAR AME Programmatic Funding Scheme Under Project A20H4b0141, A*STAR Central Research Fund, the Science and Technology Department of Sichuan Province (Grant No. 2022YFS0071 \& 2023YFG0273), and the China Scholarship Council (No. 202206240082).}

\bibliography{paper240}
\bibliographystyle{splncs04}
\newpage
\begin{appendix}
\section*{Supplementary Materials}
\setcounter{figure}{0}
\setcounter{table}{0}
\setcounter{equation}{0}
\setcounter{page}{1}
\renewcommand{\thetable}{S\arabic{table}}
\renewcommand{\thefigure}{S\arabic{figure}}
\renewcommand{\theequation}{S\arabic{equation}}

\section*{1.  Parameters selection for $w^m, w^c, w^s$}
\begin{table}[htbp]
  \centering
  \caption{The process of parameters selection.}
    \begin{tabular}{ccccccccccc}
    \toprule
    \multirow{2}[4]{*}{0.5*$\mathcal{L}_{cls}$} & \multirow{2}[4]{*}{0.4*$\mathcal{L}_{seg}$} & \multirow{2}[4]{*}{0.1*$\mathcal{L}_{mut}$} &       & \multicolumn{2}{c}{CLS} &       & \multicolumn{4}{c}{SEG} \\
\cmidrule{5-6}\cmidrule{8-11}          &       &       &       & $ACC$   & $F1 $   &       & $DI_{disc}$ & $ASSD_{disc}$ & $DI_{cup}$ & $ASSD_{cup}$ \\
\cmidrule{1-3}\cmidrule{5-6}\cmidrule{8-11}    $\checkmark$     &       &       &       & \underline{0.840}  & \underline{0.858}  &       & 0.031  & 63.992  & 0.000  & 78.304  \\
    $\checkmark$     & $\checkmark$     &       &       & 0.805  & 0.836  &       & \underline{0.771}  & \underline{1.925}  & \underline{0.805}  & \underline{3.604}  \\
    $\checkmark$     & $\checkmark$     & $\checkmark$     &       & \textbf{0.853} & \textbf{0.875} &       & \textbf{0.855} & \textbf{1.560} & \textbf{0.858} & \textbf{1.251} \\
    \bottomrule
    \end{tabular}%
  \label{tab:addlabel}%
\end{table}%

\section*{2. More visualizations}
\subsection*{2.1 Additional visual results}
\begin{figure}
\centering
\includegraphics[width=1\linewidth]{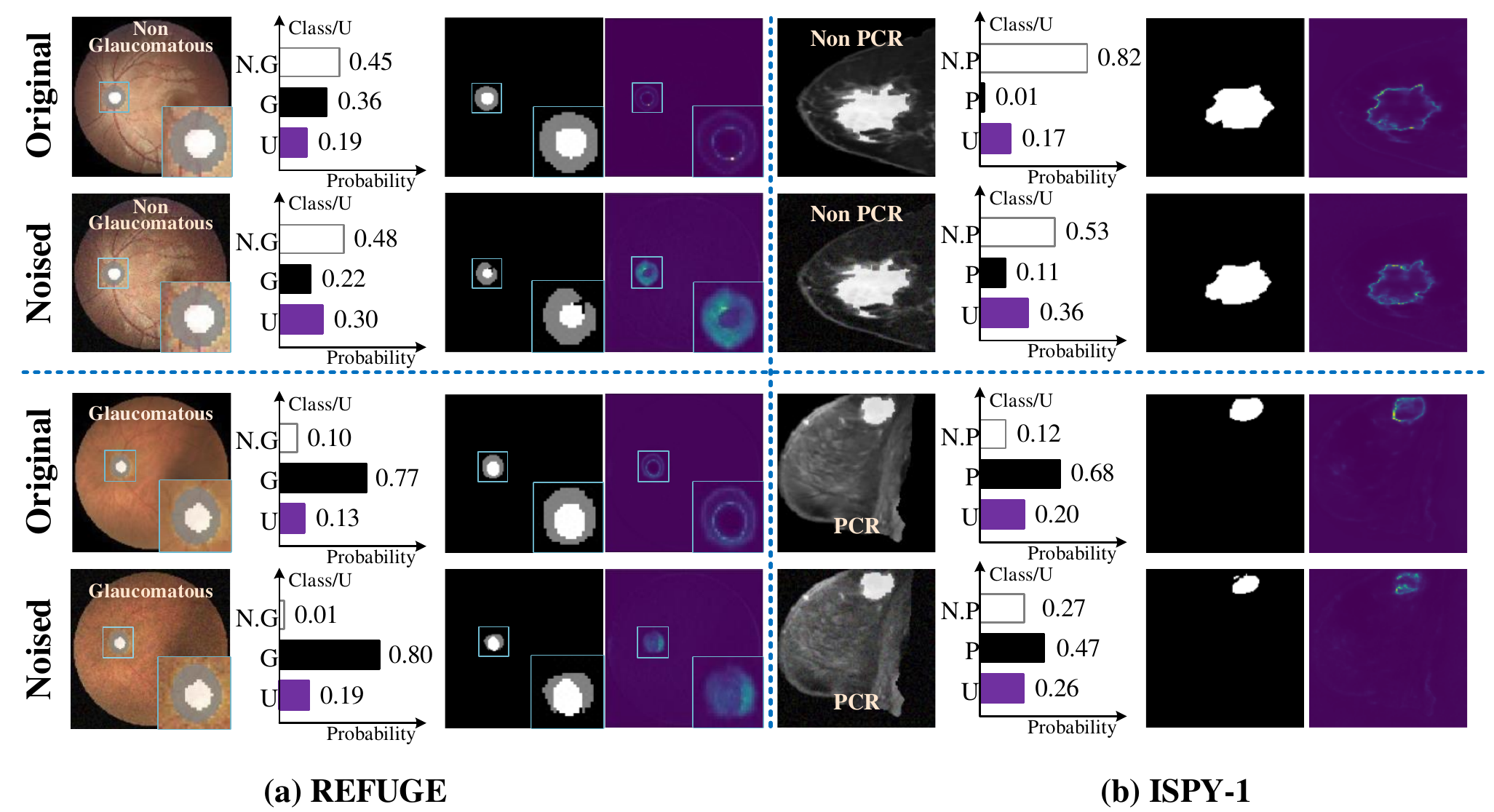}
\vskip -5pt
\caption{Four additional visual results of segmentation and classification in REFUGE and ISPY-1. Top is the original image, and bottom is the input with Gaussian noise ($\sigma = 0.05$). From left to right, input (with GT), the result of classification (belief and image-level uncertainty), the result of segmentation, pixel-wise uncertainty.}
\label{AF_2}
\end{figure}

\newpage
\subsection*{2.2 Attention maps of classification}
\vskip -20pt
\begin{figure}
\centering
\includegraphics[width=0.9\linewidth]{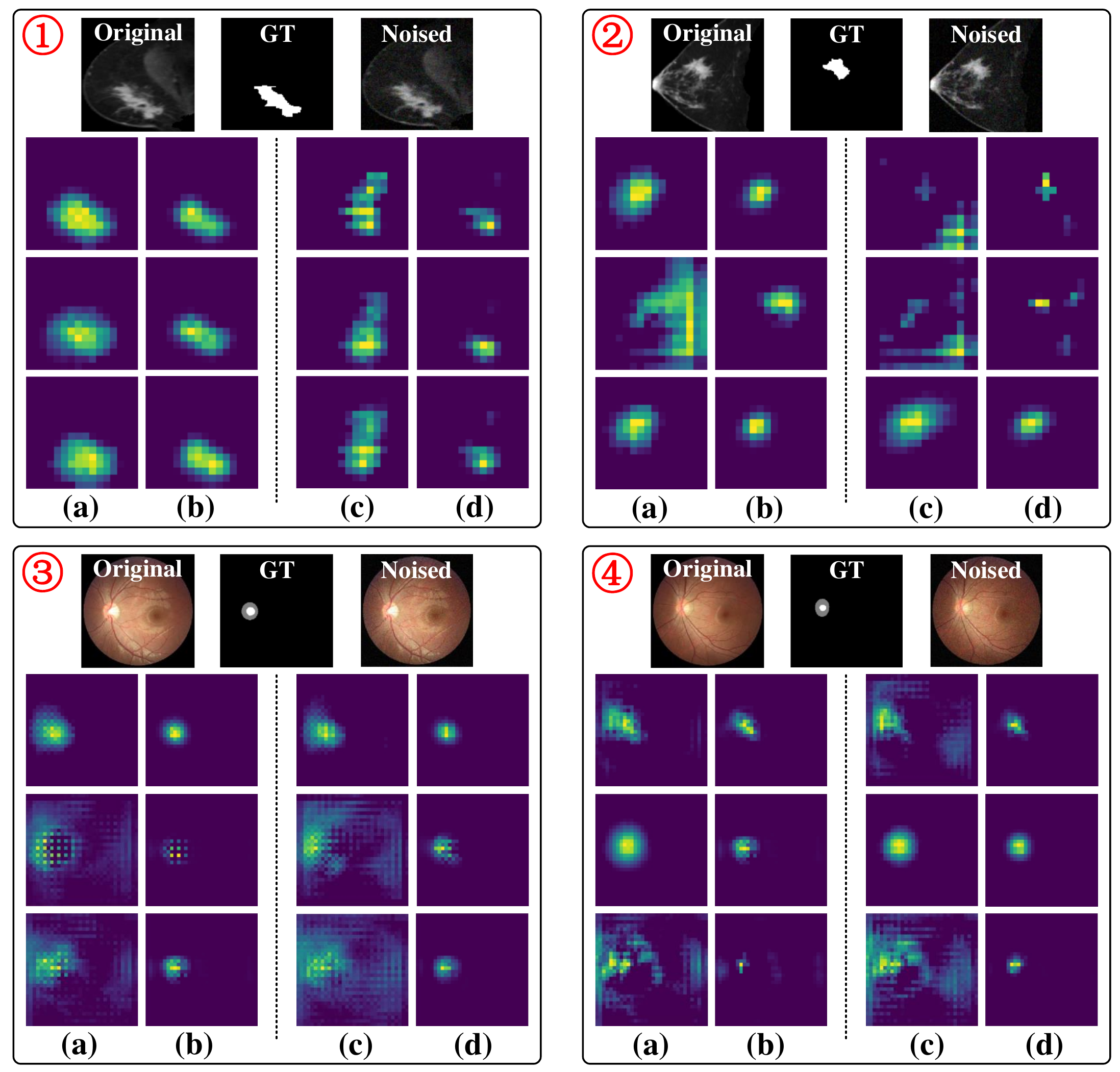}
\vskip -5pt
\caption{The attention map of classification in REFUGE and ISPY-1 whose inputs are original or noised ($\sigma = 0.05$) images. \textcolor{red}{\textcircled{1}} and  \textcolor{red}{\textcircled{2}} are in ISPY-1. \textcolor{red}{\textcircled{3}} and  \textcolor{red}{\textcircled{4}} are in REFUGE. (a), (b) represent the channel maps of $f_c^4$ and $r^c$ of the original images. (c), (d) represent the channel maps of $f_c^4$ and $r^c$ of the noised images. The three channel maps are extracted from 1024 feature maps, randomly.}
\vskip -30pt
\label{AF_1}
\end{figure}

\subsection*{2.3 Qualitative comparison with multi-task methods.}
\vskip -20pt
\begin{figure}
\centering
\includegraphics[width=\linewidth]{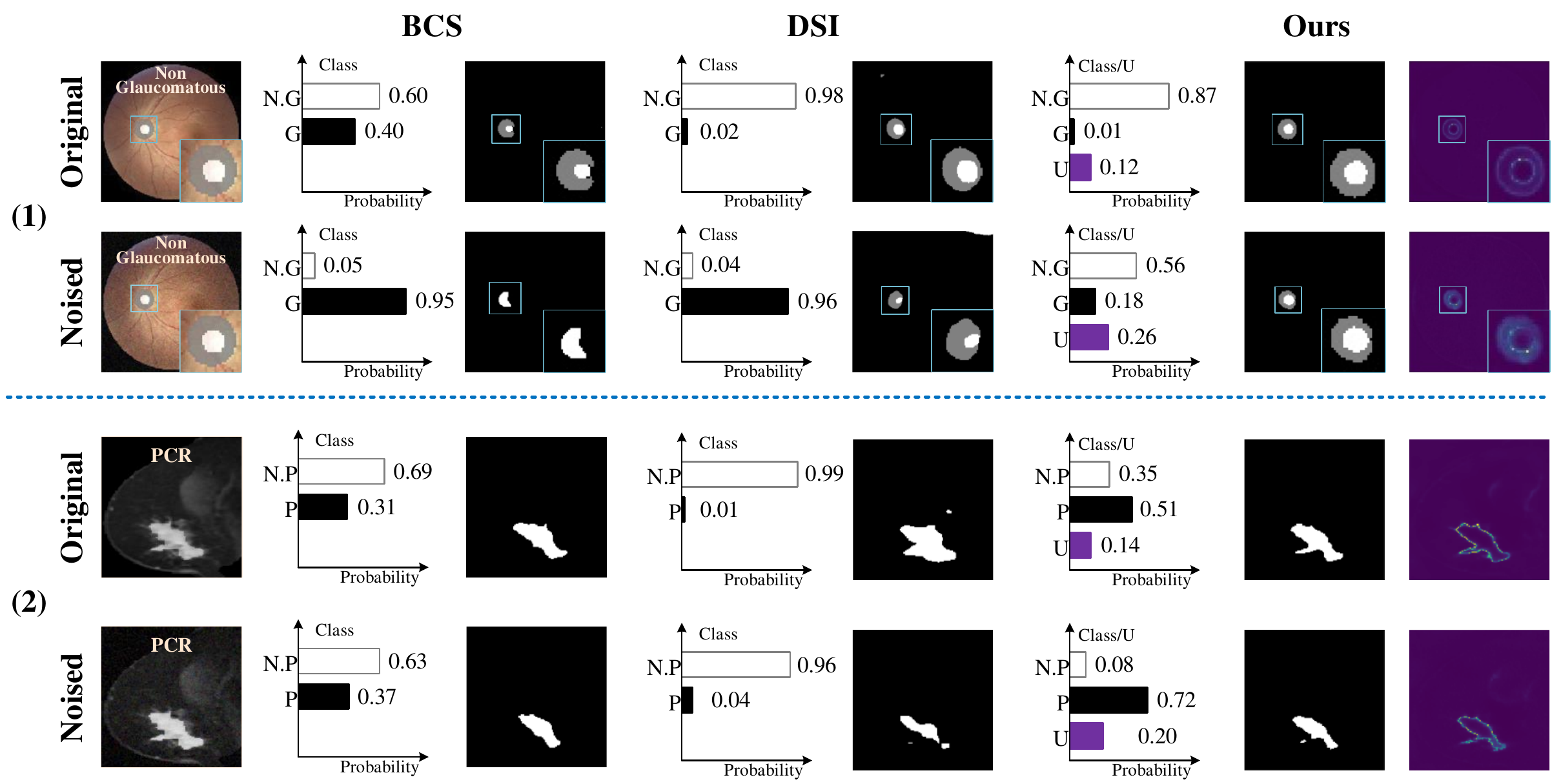}
\vskip -5pt
\caption{The visual comparisons of classification and segmentation results in two datasets ((1) is REFUGE and (2) is ISPY-1) with multi-task methods.}
\vskip -200pt
\label{AF_3}
\end{figure}
\end{appendix}
\end{document}